\documentclass[runningheads,a4paper]{llncs}
\usepackage{multirow}
\usepackage{amsmath,amssymb}

\usepackage{graphicx}
\usepackage[caption=false]{subfig}
\usepackage[table,x11names]{xcolor}

\usepackage{url}
\newcommand{\keywords}[1]{\par\addvspace\baselineskip
\noindent\keywordname\enspace\ignorespaces#1}

\usepackage{xcolor}

\begin{document}

\mainmatter  

\title{Unsupervised learning for cross-domain medical image synthesis using deformation invariant cycle consistency networks}

%
%
\author{Chengjia Wang $ ^{1,2}$ 
 \thanks{This work was supported by British Heart Foundation.}
\and Gillian Macnaught$ ^{1,2} $ 
\and Giorgos Papanastasiou$ ^{2} $ 
\and Tom MacGillivray$ ^{2} $ 
\and David Newby$ ^{1,2} $ 
}

\institute{$^{1}$BHF Centre for Cadiovascular Science, University of Edinburgh, Edinburgh, UK $^{2}$Edinburgh Imaging Facility QMRI, University of Edinburgh, Edinburgh, UK\\
\url{chengjia.wang@ed.ac.uk} 
}

%
%

\toctitle{Lecture Notes in Computer Science}
\tocauthor{Authors' Instructions}
\maketitle

\begin{abstract}
Recently, the cycle-consistent generative adversarial networks (CycleGAN) has been widely used for synthesis of multi-domain medical images. The domain-specific nonlinear deformations captured by CycleGAN make the synthesized images difficult to be used for some applications, for example, generating pseudo-CT for PET-MR attenuation correction. This paper presents a deformation-invariant CycleGAN (DicycleGAN) method using deformable convolutional layers and new cycle-consistency losses. Its robustness dealing with data that suffer from domain-specific nonlinear deformations has been evaluated through comparison experiments performed on a multi-sequence brain MR dataset and a multi-modality abdominal dataset. Our method has displayed its ability to generate synthesized data that is aligned with the source while maintaining a proper quality of signal compared to CycleGAN-generated data. The proposed model also obtained comparable performance with CycleGAN when data from the source and target domains are alignable through simple affine transformations.

\end{abstract}

\section{Introduction}
Modern clinical practices make cross-domain medical image synthesis a technology gaining in popularity. (In this paper, we use the term ``domain'' to uniformly address different imaging modalities and parametric configurations.) Image synthesis allows one to handle and impute data of missing domains in standard statistical analysis \cite{Tulder2015}, or to improve intermediate step of analysis such as registration \cite{Iglesias2013}, segmentation \cite{Roy2011} and disease classification \cite{Li2014}. Our application is to generate pseudo-CT images from multi-sequence MR data \cite{Nie2017}. The synthesized pseudo-CT images can be further used for the purpose of PET-MR attenuation correction \cite{Wagenknecht2013}.

\begin{figure}[b]
\centering
\includegraphics[width=0.95\textwidth]{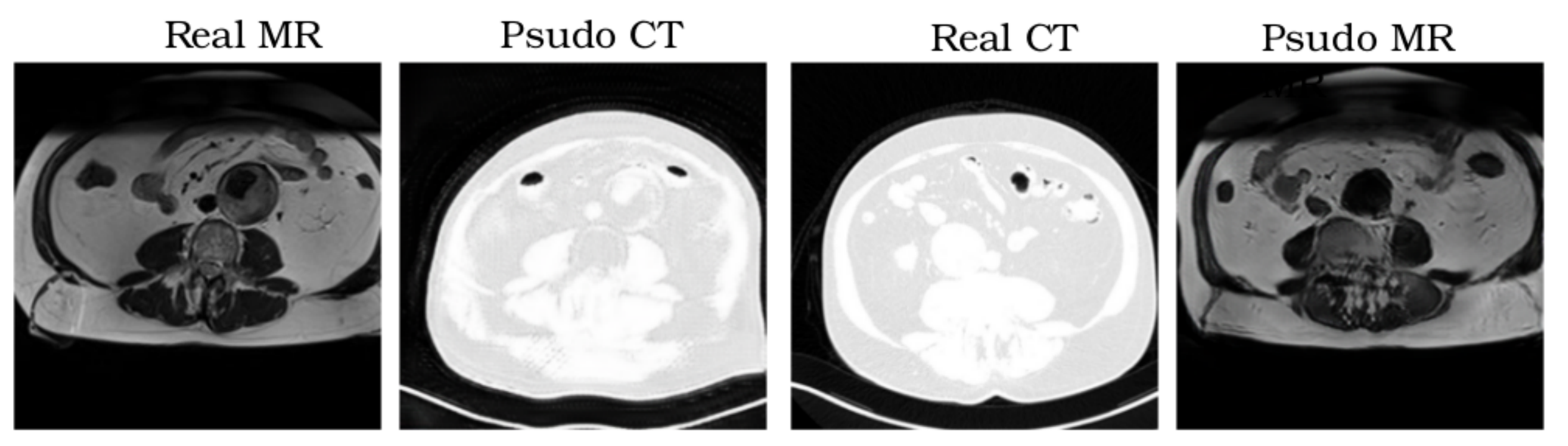}
\caption{Example of MR-CT sysnthesis using vanila CycleGAN.}
\label{fig:extreme_example}
\end{figure}

State-of-the-art methods often train a deep convolutional neural network (CNN) as image generator  following the learning procedure of the generative adversarial network (GAN) \cite{Goodfellow2014}. Many of these methods require to use aligned, or paired, datasets which is hard to obtain in practice when the data can not be aligned through an affine transformation. To deal with unpaired cross-domain data, a recent trend is to leverage CycleGAN losses \cite{Zhu2017} into the learning process to capture high-level information translatable between domains. Previous studies have shown that CycleGAN is robust to unpaired data \cite{Wolterink2017}. However, not all information encoded in CycleGAN image generators should be used due to very distinct imaging qualities and characteristics in different domains, especially different modalities. Fig. \ref{fig:extreme_example} displays a representative example of CycleGAN based cross-modality synthesis where the real CT and MR data were acquired from the same patient. It can be seen that the shape and relative positions of the scanner beds are very different. This problem can be addressed as ``domain-specific deformation''. Because the generator networks can not treat the spatial deformation and image contents separately, CycleGAN encodes this information and reproduce it in the forward pass, which causes misalignment between the source and synthesized images. For some applications, such as generating pseudo-CT for attenuation correction of PET-MR data, this domain-specific deformation should be discarded. In the mean time, the networks should keep efficient information about appearences of the same anatomy in distinct domains. A popular strategy to solve this problem is performing supervised or semi-supervised learning with an extra mission, for example, segmentation \cite{Huo2017}, but this requires collection of extra ground truth. 

In this paper, we present a deformation invariant CycleGAN (DicycleGAN) framework for cross-domain medical image synthesis. The architecture of the networks is inspired by the design of deformable convolutional network (DCN) \cite{Dai2017}. We handle the different nonlinear deformations in different domains by integrating a modified DCN structure into the image generators and propose to use normalized mutual information (NMI) in the CycleGAN loss. We evaluate the proposed network using both multi-modality abdominal aortic data and multi-sequence brain MR data. The experimental results demonstrate the ability of our method to handle highly disparate imaging domains and generate synthesized images aligned with the source data. In the mean time, the quality of the synthesized images are as good as those generated by the CycleGAN model. The main contributions of this paper include a new DicycleGAN architecture which learns deformation-invariant correspondences between domains and a new NMI-based cycleGAN loss. 

\section{Method}
A GAN framework using a image generator $ G $ to synthesize images of a target domain using images from a source domain, and a discriminator $D$ to distinguish real and synthesized images. Parameters of $G$ are optimized to confuse $D$, while $D$ is trained at the same time for better binary classification performance to classify real and synthesized data. We assumes that we have $ n^A $ images $ x^A \in \mathcal{X}^A $ from domain $\mathcal{X}^A$, and $ n^B $ images $ x^B \in \mathcal{X}^B $. To generate synthesized images of domain $ \mathcal{X}^B $ using images from $ \mathcal{X}^A $, $ G $ and $ D $ are trained in the min-max game of the GAN loss $ \mathcal{L}_{GAN} \left( G, D, \mathcal{X}^A, \mathcal{X}^B \right) $ \cite{Goodfellow2014}.  When dealing with unpaired data, the original CycleGAN framework consists of two symmetric sets of generators $G^{A\rightarrow B}$ and $G^{B\rightarrow A}$ act as mapping functions applied to a source domain, and two discrimitors $ D^B $ and $D^A$ to distinguish real and synthesized data for a target domain. The \textit{cycle consistency} loss  $\mathcal{L}_{cyc} \left(G^{A\rightarrow B}, D^A, G^{B\rightarrow A}, D^B, \mathcal{X}^A, \mathcal{X}^B \right)$, or $ \mathcal{L}^{A, B}_{cyc} $, is used to keep the cycle-consistency between the two sets of networks. The loss of the whole CycleGAN framework $\mathcal{L}_{CycleGAN} = \mathcal{L}^{A\rightarrow B}_{GAN} + \mathcal{L}^{B\rightarrow A}_{GAN} + \lambda_{cyc} \mathcal{L}^{A,B}_{cyc}$. (In this paper we use the short expression $ \mathcal{L}^{\mathcal{A}\rightarrow \mathcal{B}}_{GAN} $ to denote GAN loss $ \mathcal{L}_{GAN}(G^{A\rightarrow B}, D^B, \mathcal{X}^A, \mathcal{X}^B) $). The image generator in the CycleGAN contains an input convolutional block, two down-sampling convolutional layers, followed by a few resnet blocks or a Unet structure, and two up-sampling transpose convolutional blocks before the last two convolutional blocks.  

\subsubsection{DicycleGAN Architecture}
\begin{figure}[!t]
\centering
\includegraphics[width=0.9\linewidth]{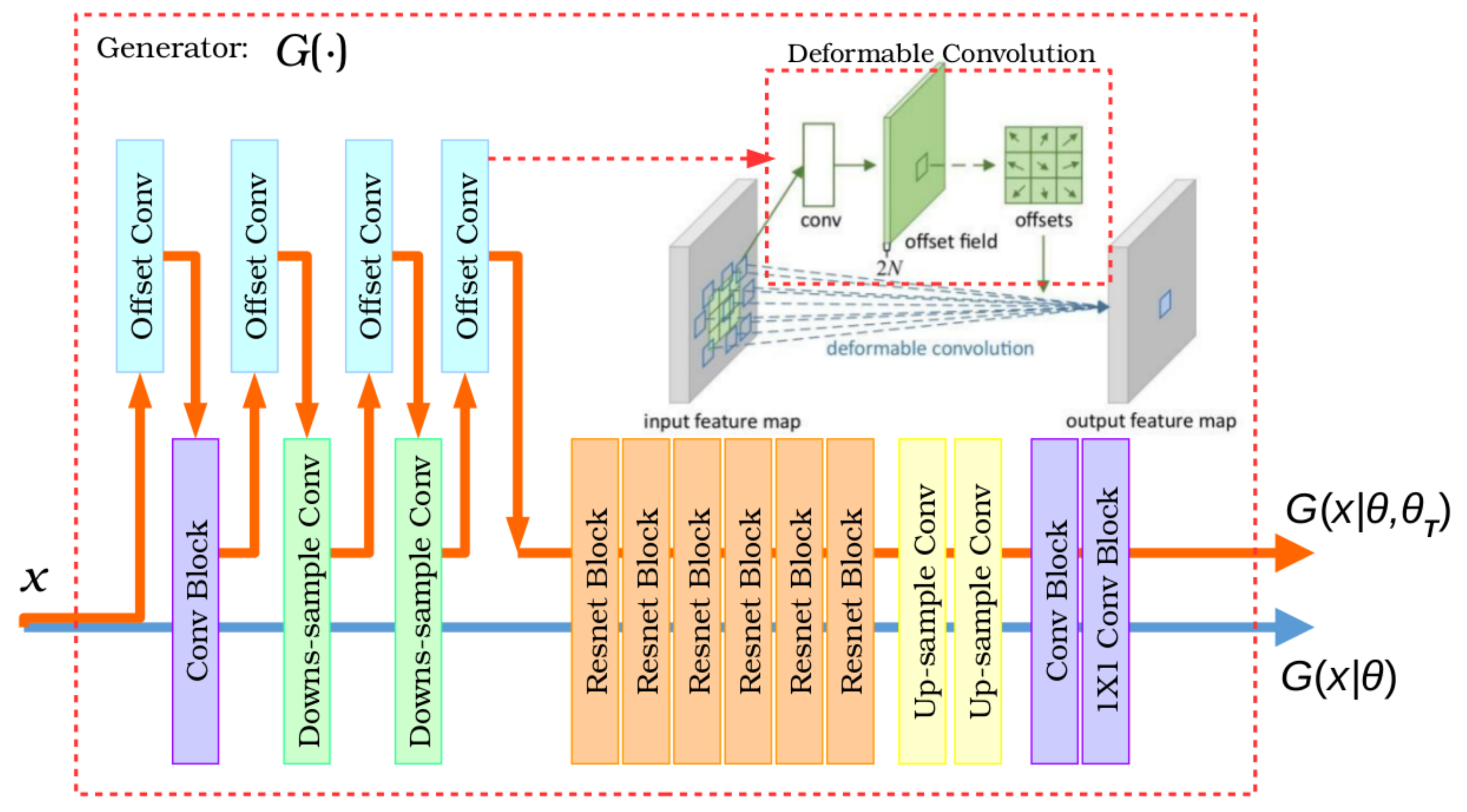}
\caption{Architecture of the proposed image generator $ G(\cdot) $. Each input image $ x $ generates a deformed output $ G(x|\theta, \theta_T) $ and an undeformed output $ G(x|\theta) $ through two forward passes shown in red and blue. Demonstration of deformable convolution is obtained from \cite{Dai2017}. (best viewed in color)}
\label{fig:DicycleGAN_G}
\end{figure}

In order to capture deformation-invariant information between domains, we introduce a modified DCN architecture into the image generators of CycleGAN, as shown in Fig. \ref{fig:DicycleGAN_G}. The deformable convolution can be viewed as an atrous convolution kernel with trainable dilation rates and reinterpolated input feature map \cite{Dai2017}. The spatial offsets of each point in the feature map is learned through a conventional convolution operation, followed by another convolution layer. This leads to a ``Lasagne'' structure consist of interleaved ``offset convolution'' and conventional convolution operations. We adopt this structure to the generators by inserting an offset convolutional operation (displayed in cyan in Fig. \ref{fig:DicycleGAN_G}) in front of the input convolutional block, down-sample convolutional blocks and the first resnet blocks. Note that this ``offset'' convolution only affects the interpolation of the input feature map rather than providing a real convolution result. Let $ \theta _T $ denote the learnable parameters in the``offset'' convolutional layers, and $ \theta $ the rest parameters in image generator $ G $. When training $ G $, each input image $ x $ generates two output images: deformed output image $ G_T(x) = G\left( x|\theta, \theta _T \right) $ and undeformed image $ G(x) = G\left( x| \theta \right) $. The red and blue arrows in Fig. \ref{fig:DicycleGAN_G} indicate the computation flows for generating $ G_T(x) $ and $G(x)$ in the forward passes. $ G_T(x) $ is then taken by the corresponding discriminator $ D $ for calculation of GAN losses, and $ G(x) $ is expected to be aligned with $ x $.

\subsubsection{DicycleGAN Loss}
DicycleGAN loss contains traditional GAN loss following the implementation of CycleGAN \cite{Zhu2017}, but also includes an image alignment loss and a new cycle consistency loss.  For the GAN loss $ L_{GAN}^{A \rightarrow B}$, the image generator $G^{A \rightarrow B}$ is trained to minimize $ \left( D^B \left( G_T^{A\rightarrow B} \left( x^A \right) \right) -1 \right)^2 $ and $ D^B $ is trained to minimize  $ \left( D^B(x^B) -1 \right)^2 + D^B \left( G_T^{A\rightarrow B}(x^A) \right)^2 $. The same formulation is used to calculate $ \mathcal{L}^{B\rightarrow A}_{GAN} $ defined on $ G^{B\rightarrow A}$ and $ D^A $. Note that the GAN loss is calculated based on the deformed synthesied images. As the undeformed outputs of generators are expected to be aligned with the input images, we propose to use a information loss based on normalized mutual information (NMI). NMI is a popular metric used for image registration. It varies between 0 and 1 indicating alignment of two clustered images \cite{Vinh2010}. The image alignment loss is defined as:
\begin{equation}
\mathcal{L}_{align}^{A, B} = 2 -NMI\left( x^A, G^{A \rightarrow B}\left( x^A \right) \right) -NMI \left( x^B, G^{B\rightarrow A} \left( x^B \right) \right).
\label{equ:loss_align}
\end{equation}
Based on the proposed design of image generators, the cycle two types of cycle consistency losses. The undeformed cycle consistency loss is defined as:
\begin{equation}
\mathcal{L}_{cyc}^{A, B} = \| G^{B\rightarrow A}\left(G^{A\rightarrow B} \left( x^A \right) \right) - x^A \|_{1} + \| G^{A\rightarrow B}\left( G^{B\rightarrow A} \left( x^B \right) \right) - x^B
 \|_{1}.
\end{equation}
Beside $ \mathcal{L}_{cyc} $, the deformation applied to the synthesized image should be also cycle consistent. Here we defined a deformation-invariant cycle consistency loss:
\begin{equation}
\mathcal{L}_{dicyc}^{A, B} = \| G_T^{B\rightarrow A}\left(G_T^{A\rightarrow B} \left( x^A \right) \right) - x^A \|_{1} + \| G_T^{A\rightarrow B}\left( G_T^{B\rightarrow A} \left( x^B \right) \right) - x^B
 \|_{1}. 
\end{equation}
To perform image synthesis between domains $ \mathcal{X}^{A} $ and $ \mathcal{X}^B $, we use the deformed output images $ G_{T}^{A\rightarrow B} $ and $ G_{T}^{B\rightarrow A} $ to calculate the GAN loss.  The full loss of DicycleGAN is:
\begin{equation}
\mathcal{L}_{DicycleGAN} = \mathcal{L}_{GAN}^{A\rightarrow B} + \mathcal{L}_{GAN}^{B \rightarrow A} +  \lambda_{align} \mathcal{L}_{align}^{A, B} + \lambda_{cyc} \mathcal{L}_{cyc}^{A, B} + \lambda_{dicyc} \mathcal{L}_{dicyc}^{A, B}.
\end{equation} 

\begin{figure}[!t]
\subfloat[\label{subfig-1:loss1}]{%
       \includegraphics[width=0.5\linewidth]{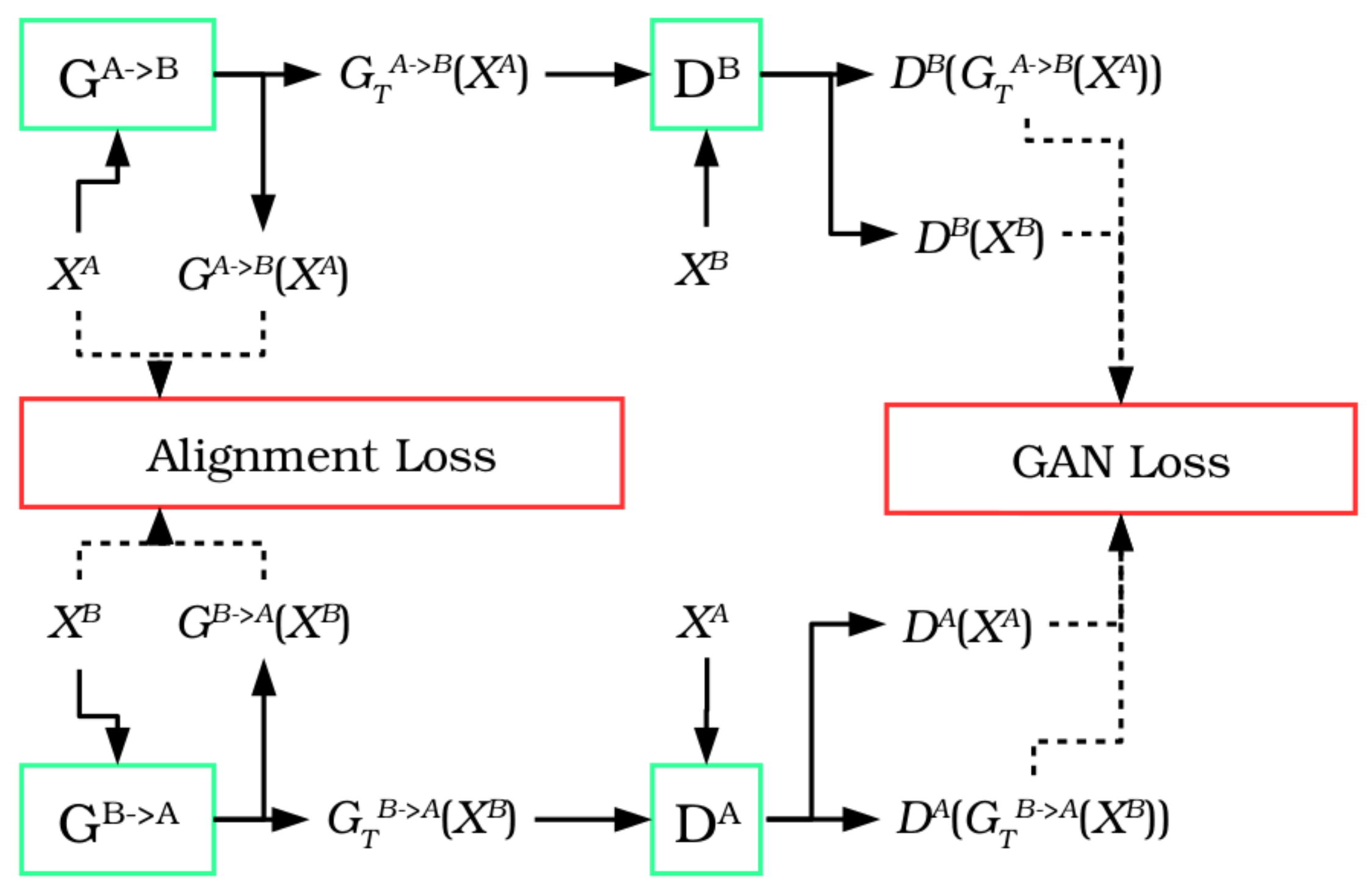}
     }
     \hfill
     \subfloat[\label{subfig-2:loss2}]{%
       \includegraphics[width=0.5\linewidth]{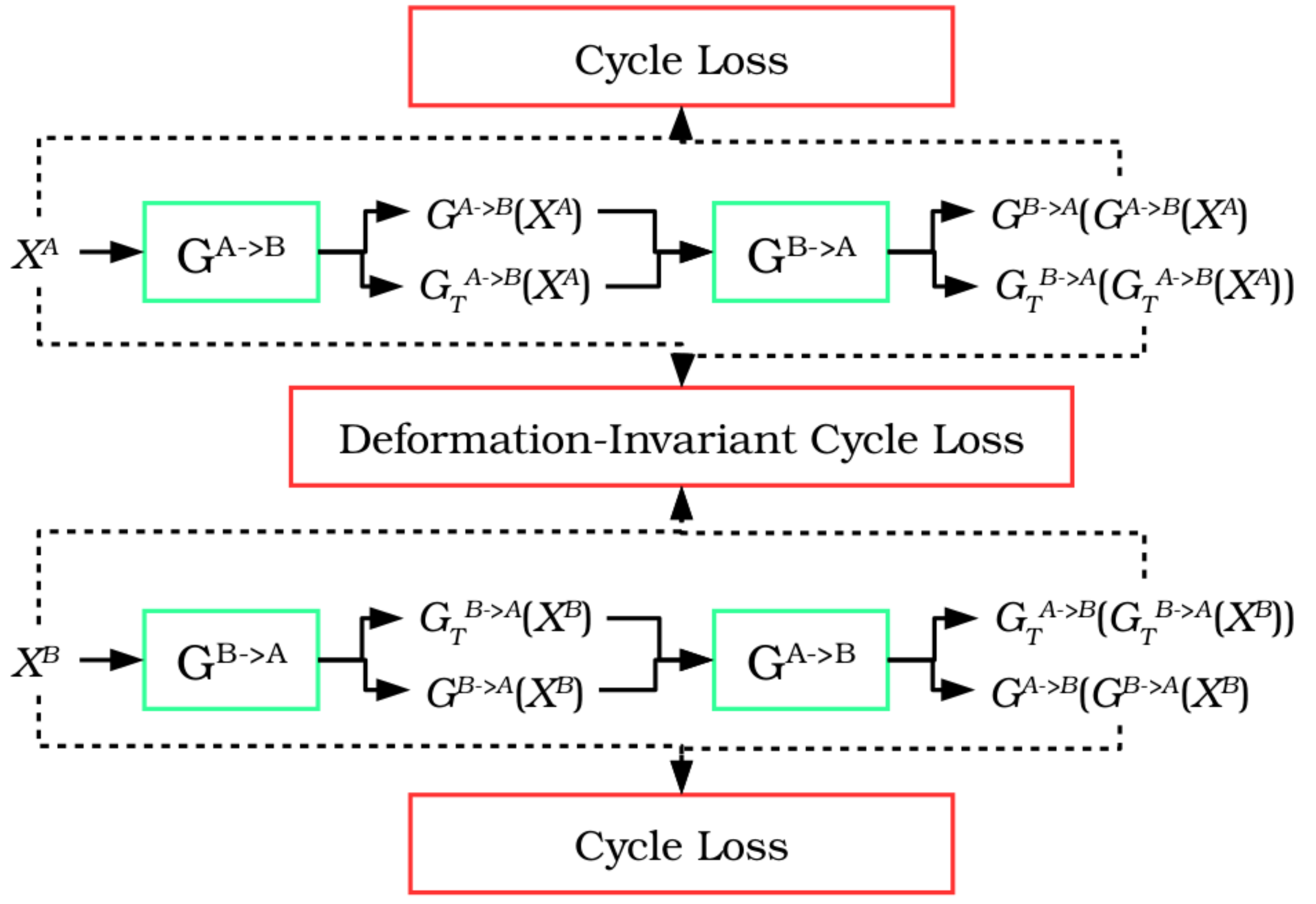}
     }
     \caption{Calculation of losses in DicycleGAN. \ref{subfig-1:loss1} shows GAN and image alignment losses: the undeformed output of the image generators are used for alignment losses, and the deformed outputs for GAN losses. \ref{subfig-2:loss2} shows the Cycle consistency losses.}
     \label{fig:losses}
\end{figure}

Fig. \ref{fig:losses} provides a demonstration of computing the all the losses discussed above using outputs of the corresponding DicycleGAN generators and discriminators.

\section{Experiments}
\subsubsection{Evaluation Metrics}
The most widely used quantative evaluation metrics for cross-domain image synthesis are: mean squared error (MSE), peak signal-to-noise ratio (PSNR) and structural similarity index (SSIM). Given a volume $ x^A $ and a target volume $ x^B $, the MSE is computed as: $ \frac{1}{N} \sum^{N}_{1} \left( x^B - G^{A\rightarrow B}(x^A)  \right)^2 $, where $N$ is number of voxels in the volume. PSNR is calculated as: $10 \log _{10} \frac{\max_B^2}{MSE}$. SSIM is computed as: $ \frac{(2\mu_A \mu_B + c_1)(2\delta_{AB} + c2)}{(\mu_{A}^2 + \mu_{B}^2 +c_1)(\delta_{A}^2+ \delta+{B}^2 + c2)} $, where $ \mu $ and $ \delta^2 $ are mean and variance of a volume, and $\delta_{AB}$ is the covariance between $ x^A $ and $x^B$. $ c_1$ and $c_2$ are two variables to stabilize the division with weak denominator \cite{Larkin2015}.

\subsubsection{Datasets}
We use the Information eXtraction from Images (IXI) dataset \footnote{http://brain-development.org/ixi-dataset/} which provides co-registered multi-sequence skull-stripped MR images collected from multiple sites. Due to the limited storage space, here we selected 66 proton density (PD-) and T2-weighted volumes, each volume contains 116 to 130 2D slices. We use 38 pairs for training and 28 pairs for evaluation of synthesis results. Our image generators take 2D axial-plane slices of the volumes as inputs. During the training phase, we resample all volumes to a resolution of $ 1.8 \times 1.8 \times 1.8 mm^3/voxel $, then crop the volumes to $ 128 \times 128 $ pixel images. As the generators in both CycleGAN and DicycleGAN are fully convolutional, the predictions are performed on full size images. All the images are normalized with their mean and standard deviation. We also used a private dataset contains 40 multi-modality abdominal T2*-weighted images and CT images collected from 20 patients with abdominal aortic aneurysm (AAA) in a clinical trial. All images are resampled to a resolution of $ 1.56 \times 1.56 \times 5 mm^3 /voxel $, and the axial-plane slices trimmed to $ 192 \times 192 $ pixels. It is difficult to non-rigidly register whole abdominal images to calculate the evaluation metrices, but the aorta can be rigidly aligned to assess the performance of image synthesis. The anatomy of the aorta have previously been co-registered and segmented by 4 clinical researchers. 

\subsubsection{Implementation Details}
We used image generators with 9 Resnet blocks. All parameters of, or inherit from, vanilla CycleGAN are taken from the original paper. For the DicycleGAN, we set $ \lambda_{cyc} = \lambda_{dicyc} = 10 $ and $ \lambda_{align} = 0.9 $. The models were trained with Adam optimizer \cite{Kingma2014} with a fixed learning rate of $0.0002$ for the first 100 epochs, followed by 100 epochs with linearly decreasing learning rate. Here we apply a simple early stop strategy: in the first 100 epochs, when $ \mathcal{L}_{DicycleGAN} $ stops decreasing for 10 epochs, the training will move to the learning rate decaying stage; this tolerance is set to 20 epochs in the second 100 epochs.

\subsubsection{Experiments Setup}
In order to quantitatively evaluate robustness of our model to the domain-specific local distortion, we applied an aribitrary non-linear distortion to the T2-weighted images of IXI. Synthesis experiments were performed between the PD-weighted data and undeformed, as well as the deformed T2-weighted data. When using deformed T2-weighted images, the ground truth were generated by applying the same nonlinear deformation to the source PD-weighted images. We trained the CycleGAN and DicycleGAN using unpaired, randomly selected slices. The training images were augmented using aggresive flips, rotations, shearing and translations so that CycleGAN can be robust. In the test stage, the three metrics introduced above were computed. For our private dataset, the metrics were computed within the segmented aortic anatomy excluding any other imaged objects because all the three metrics require to be calculated on aligned images. 

\section{Results}
\begin{figure}[!t]
\centering
\includegraphics[width=0.95\linewidth]{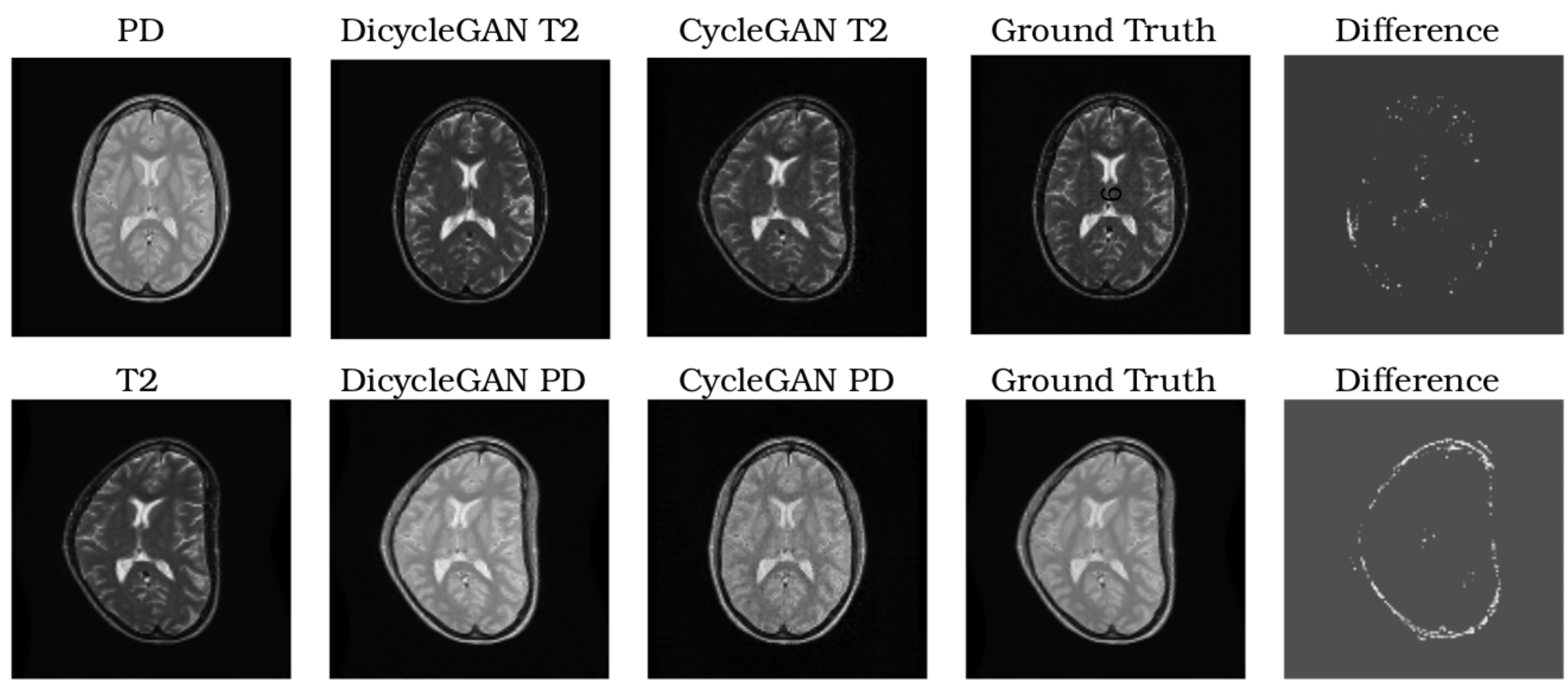}
\caption{Example of synthesized images generated by CycleGAN and DicycleGAN, compared to the ground truths. The ground truth of the deformed source image is generated by applying the same arbitrary deformation to the original target image.}
\label{fig:Brain_res}
\end{figure}

Table \ref{tab:Brain_results} and \ref{tab:Brain_resultsd} present the PD-T2 co-synthesis results using undeformed and deformed T2-weighted images. In addition, Fig. \ref{fig:Brain_res} provides an example showing the synthesized images generated by CycleGAN and DicycleGAN. CycleGAN encoded the simulated domain-specific deformation, whether applied to source or target domain, and combined this deformation into the synthesized images. This leads to misalignment of source and synthesized images. The quantitative results show that our DicycleGAN model produced comparable results with CycleGAN when there is no domain-specific distortions, but it achieved remarkable performance gain when the source and target images can not be aligned through an affine transformation. This is because of the deformed synthesized images generated by CycleGAN which lead to severe misalignments between the source and synthesized images. 

The cross-modality synthesis results are shown in Tabel \ref{tab:mars_results}. The discripancy between the two imaging modalities can be shown by the different relative positions between the imaged objects and the beds. CycleGAN encoded this information in the image generators as shown earlier in Fig. \ref{fig:extreme_example}.

\begin{table*}[!t]
\caption{Synthesis results of IXI dataset using undeformed T2 images.}
\begin{center}
\begin{tabular}{|l|c|ccc|}
\hline
\multicolumn{1}{|c|}{Experiment}& Model & MSE & PSNR & SSIM \\
\hline
\multirow{2}{*}{\textbf{PD to T2}}& Cycle &0.186 (0.08) & \textbf{27.35 (1.69)} & 0.854 (0.03) \\
\cline{2-5}
& Dicycle &\textbf{0.183 (0.09)} & 26.49 (1.62) & \textbf{0.871 (0.03)} \\
\hline
\multirow{2}{*}{\textbf{T2 to PD}}& Cycle & \textbf{0.134 (0.02)}& \textbf{29.68 (1.61)}& \textbf{0.892 (0.03)} \\
\cline{2-5}
& Dicycle & 0.146 (0.03) & 28.85 (1.59) & 0.883 (0.02)  \\
\hline
\end{tabular}
\end{center}
\label{tab:Brain_results}
\end{table*}

\begin{table*}[!t]
\caption{Synthesis results of IXI dataset using deformed T2 images.}
\begin{center}
\begin{tabular}{|l|c|ccc|}
\hline
\multicolumn{1}{|c|}{Experiment}& Model & MSE & PSNR & SSIM\\
\hline
\multirow{2}{*}{\textbf{PD to T2}}& Cycle &0.586 (0.25) & 19.52 (1.62) & 0.6081 (0.12) \\
\cline{2-5}
& Dicycle &\textbf{0.145 (0.02)} & \textbf{22.32 (1.29)} &  \textbf{0.7842 (0.03)} \\
\hline
\multirow{2}{*}{\textbf{T2 to PD}}& Cycle & 0.561 (0.22)& 19.42 (1.61) & 0.6001 (0.11) \\
\cline{2-5}
& Dicycle & \textbf{0.141 (0.02)} & \textbf{22.86 (1.31)} & \textbf{0.7714 (0.02)} \\
\hline
\end{tabular}
\end{center}
\label{tab:Brain_resultsd}
\end{table*}

\begin{table*}[!t]
\caption{Multi-modality synthesis results using private dataset.}
\begin{center}
\begin{tabular}{|l|c|ccc|}
\hline
\multicolumn{1}{|c|}{Experiment}& Model & MSE & PSNR & SSIM\\
\hline
\multirow{2}{*}{\textbf{T2* to CT}}& Cycle &0.516 (0.19) & 18.32 (1.82) & 0.5716 (0.15) \\
\cline{2-5}
& Dicycle &\textbf{0.287 (0.11)} & \textbf{23.71 (1.17)} &  \textbf{0.7122 (0.03)} \\
\hline
\multirow{2}{*}{\textbf{CT to T2*}}& Cycle & 0.521 (0.22)& 19.12 (1.60) & 0.5818 (0.12) \\
\cline{2-5}
& Dicycle & \textbf{0.299 (0.08)} & \textbf{22.66 (1.11)} & \textbf{0.7556 (0.02)} \\
\hline
\end{tabular}
\end{center}
\label{tab:mars_results}
\end{table*}

\section{Conclusion and Discussion}
We propose a new method for cross-domain medical image synthesis, called DicycleGAN. Compared to the vanilla CycleGAN method, we integrate DCN layers into the image generators and reinforce the training process with deformation-invariant cycle consistency loss and NMI-based alignment loss. Results obtained from both multi-sequence MR dataset and our private multi-modality abdominal dataset shows that our model achieved comparable performance with CycleGAN when the source and target data can be aligned with an affine transformation. Our model achieved obvious performance gain compared to CycleGAN when there are domain-specific nonlinear deformations.  A possible future application of DicycleGAN is multi-modal image registration. 

%
%
%
%
%
%
%

\bibliographystyle{splncs}
\bibliography{ref_sysnthesis}

\begin{thebibliography}{10}

\bibitem{Tulder2015}
van Tulder, G., de~Bruijne, M.:
\newblock Why does synthesized data improve multi-sequence classification?
\newblock In: International Conference on Medical Image Computing and
  Computer-Assisted Intervention, Springer (2015)  531--538

\bibitem{Iglesias2013}
Iglesias, J.E., Konukoglu, E., Zikic, D., Glocker, B., Van~Leemput, K., Fischl,
  B.:
\newblock Is synthesizing mri contrast useful for inter-modality analysis?
\newblock In: International Conference on Medical Image Computing and
  Computer-Assisted Intervention, Springer (2013)  631--638

\bibitem{Roy2011}
Roy, S., Carass, A., Prince, J.:
\newblock A compressed sensing approach for mr tissue contrast synthesis.
\newblock In: Biennial International Conference on Information Processing in
  Medical Imaging, Springer (2011)  371--383

\bibitem{Li2014}
Li, R., Zhang, W., Suk, H.I., Wang, L., Li, J., Shen, D., Ji, S.:
\newblock Deep learning based imaging data completion for improved brain
  disease diagnosis.
\newblock In: International Conference on Medical Image Computing and
  Computer-Assisted Intervention, Springer (2014)  305--312

\bibitem{Nie2017}
Nie, D., Trullo, R., Lian, J., Petitjean, C., Ruan, S., Wang, Q., Shen, D.:
\newblock Medical image synthesis with context-aware generative adversarial
  networks.
\newblock In: International Conference on Medical Image Computing and
  Computer-Assisted Intervention, Springer (2017)  417--425

\bibitem{Wagenknecht2013}
Wagenknecht, G., Kaiser, H.J., Mottaghy, F.M., Herzog, H.:
\newblock Mri for attenuation correction in pet: methods and challenges.
\newblock Magnetic resonance materials in physics, biology and medicine
  \textbf{26}(1) (2013)  99--113

\bibitem{Goodfellow2014}
Goodfellow, I., Pouget-Abadie, J., Mirza, M., Xu, B., Warde-Farley, D., Ozair,
  S., Courville, A., Bengio, Y.:
\newblock Generative adversarial nets.
\newblock In: Advances in neural information processing systems. (2014)
  2672--2680

\bibitem{Zhu2017}
Zhu, J.Y., Park, T., Isola, P., Efros, A.A.:
\newblock Unpaired image-to-image translation using cycle-consistent
  adversarial networks.
\newblock arXiv preprint arXiv:1703.10593 (2017)

\bibitem{Wolterink2017}
Wolterink, J.M., Dinkla, A.M., Savenije, M.H., Seevinck, P.R., van~den Berg,
  C.A., I{\v{s}}gum, I.:
\newblock Deep mr to ct synthesis using unpaired data.
\newblock In: International Workshop on Simulation and Synthesis in Medical
  Imaging, Springer (2017)  14--23

\bibitem{Huo2017}
Huo, Y., Xu, Z., Bao, S., Assad, A., Abramson, R.G., Landman, B.A.:
\newblock Adversarial synthesis learning enables segmentation without target
  modality ground truth.
\newblock arXiv preprint arXiv:1712.07695 (2017)

\bibitem{Dai2017}
Dai, J., Qi, H., Xiong, Y., Li, Y., Zhang, G., Hu, H., Wei, Y.:
\newblock Deformable convolutional networks.
\newblock CoRR, abs/1703.06211 \textbf{1}(2) (2017) ~3

\bibitem{Vinh2010}
Vinh, N.X., Epps, J., Bailey, J.:
\newblock Information theoretic measures for clusterings comparison: Variants,
  properties, normalization and correction for chance.
\newblock Journal of Machine Learning Research \textbf{11}(Oct) (2010)
  2837--2854

\bibitem{Larkin2015}
Larkin, K.G.:
\newblock Structural similarity index ssimplified.
\newblock (2015)

\bibitem{Kingma2014}
Kingma, D.P., Ba, J.:
\newblock Adam: A method for stochastic optimization.
\newblock arXiv preprint arXiv:1412.6980 (2014)

\end{thebibliography}

\end{document}